\documentclass{article}
\usepackage{ijcai24}

\usepackage{times}
\usepackage{soul}
\usepackage{url}
\usepackage[hidelinks]{hyperref}
\usepackage[utf8]{inputenc}
\usepackage[small]{caption}
\usepackage{graphicx}
\usepackage{amsmath}
\usepackage{amsthm}
\usepackage{amssymb}
\usepackage{xcolor}
\usepackage{booktabs}
\usepackage{algorithm}
\usepackage{multirow}

\usepackage{mathtools}
\newcommand{\vect}[1]{\mbox{\boldmath$#1$}}
\definecolor{lightred}{rgb}{1,0.8,0.8} 
\definecolor{lightgreen}{rgb}{0.8,1,0.8}
\definecolor{lightblue}{rgb}{0.88,0.96,1}

\usepackage{algorithmicx}
\usepackage{algorithm}
\usepackage{algpseudocode}
\usepackage{wrapfig}
\usepackage{xcolor,colortbl}
\usepackage{siunitx} 

\usepackage[switch]{lineno}

\newtheorem{example}{Example}

\title{Redefining Contributions: Shapley-Driven Federated Learning} 

\author{
Nurbek Tastan\and
Samar Fares\and
Toluwani Aremu\and
Samuel Horvath\and
Karthik Nandakumar\\
\affiliations
Mohamed bin Zayed University of Artificial Intelligence (MBZUAI), Abu Dhabi, UAE \\
\emails
\{nurbek.tastan, samar.fares, toluwani.aremu, samuel.horvath, karthik.nandakumar\}@mbzuai.ac.ae
}

\begin{document}

\maketitle

\begin{abstract}
    Federated learning (FL) has emerged as a pivotal approach in machine learning, enabling multiple participants to collaboratively train a global model without sharing raw data. While FL finds applications in various domains such as healthcare and finance, it is challenging to ensure global model convergence when participants do not contribute equally and/or honestly. To overcome this challenge, principled mechanisms are required to evaluate the contributions made by individual participants in the FL setting. Existing solutions for contribution assessment rely on general accuracy evaluation, often failing to capture nuanced dynamics and class-specific influences. This paper proposes a novel contribution assessment method called ShapFed for fine-grained evaluation of participant contributions in FL. Our approach uses Shapley values from cooperative game theory to provide a granular understanding of class-specific influences. Based on ShapFed, we introduce a weighted aggregation method called ShapFed-WA, which outperforms conventional federated averaging, especially in class-imbalanced scenarios. Personalizing participant updates based on their contributions further enhances collaborative fairness by delivering differentiated models commensurate with the participant contributions. Experiments on CIFAR-10, Chest X-Ray, and Fed-ISIC2019 datasets demonstrate the effectiveness of our approach in improving utility, efficiency, and fairness in FL systems. The code can be found at \href{https://github.com/tnurbek/shapfed}{https://github.com/tnurbek/shapfed}. 
\end{abstract}

\section{Introduction}
Federated Learning (FL) is a machine learning (ML) paradigm that enables training powerful models while respecting data privacy and decentralization. In traditional ML, data centralization poses significant privacy concerns and logistical hurdles. FL, on the other hand, flips this paradigm by allowing multiple participants or edge devices to collaborate without sharing their raw data \cite{mcmahan2017communication,li2020federated,tastan2023capride,wei2020federated}. Since only model updates are exchanged in FL, the privacy of local data sources is preserved. This approach has found applications in domains such as healthcare and finance.

In a typical FL environment, all the participants are assumed to collaborate honestly and contribute equally. The convergence and utility of a global model in FL can be hindered when this assumption is not met. When some participants intentionally or unintentionally introduce biases or skewed data distributions into the training process, it negatively impacts the overall model's performance. To address this issue, current solutions \cite{jiang2023fair,9589136,shi2022fedfaim,siomos2023contribution,wang2020principled,xu2021gradient} primarily rely on evaluating each participant's individual accuracy on an auxiliary validation set to assess their contributions to the system. 

However, these solutions often fail to detect subtle but crucial variations in how each party affects the model's performance, especially in scenarios where imbalanced data or class disparities exist. Furthermore, they fail to capture the nuanced dynamics of participant influence on specific class predictions. As a result, there is a pressing need for more refined evaluation methods that go beyond general accuracy assessment and provide a granular understanding of participant contributions to class-specific model performance, ensuring the fairness and effectiveness of FL systems. 

\begin{figure*}[t]
    \centering
    \includegraphics[width=\linewidth]{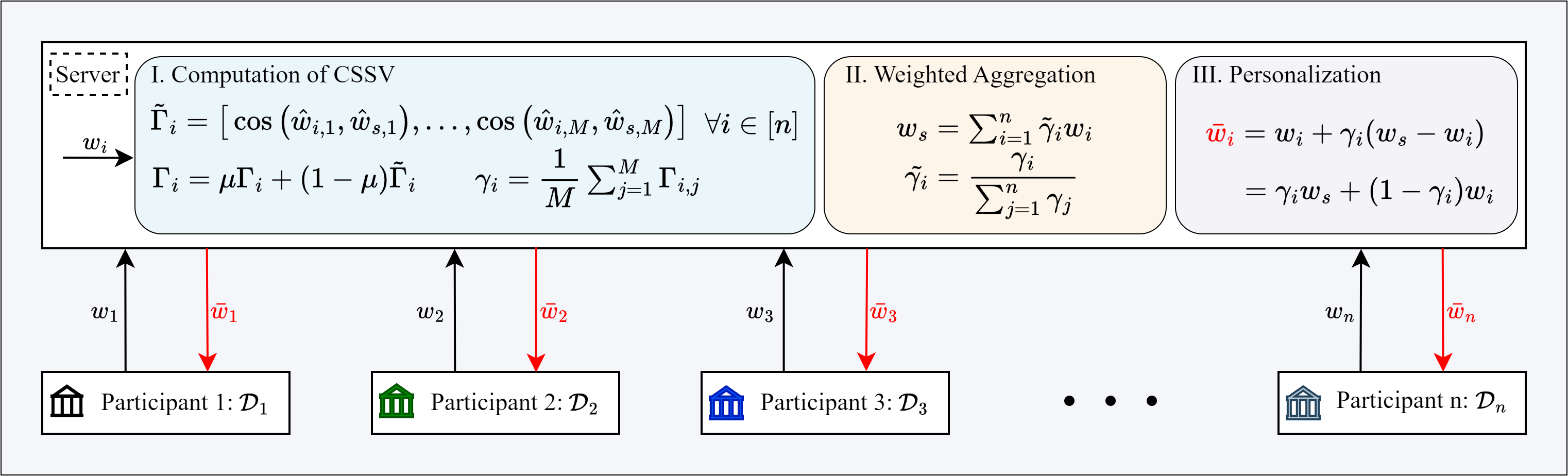}
    \caption{\textbf{Overview of our proposed} \textbf{ShapFed} \textbf{algorithm:} Each participant $i$ transmits their locally computed iterates $w_i$ to the server. The server then, (i) computes class-specific Shapley values (CSSVs) using the last layer parameters (gradients) $\hat{w}$ (as illustrated in Figure \ref{fig: network-illustration}), (ii) aggregates the weights by employing normalized contribution assessment values $\tilde{\gamma}_i$ for each participant $i$, and (iii) broadcasts the personalized weights $\bar{w}_i$ to each participant, using their individual, not-normalized contribution values $\gamma_i$. } 
    \label{fig:enter-label}
\end{figure*}

In this work, we present a novel approach to appropriately value participant contributions within the context of FL and subsequently allocate rewards. The proposed method employs Shapley values (SVs) from cooperative game theory, integrating them with class-specific performance metrics. Existing contribution assessment methods \cite{xu2021gradient,lyu2020collaborative} utilize gradients from each participant and compute the directional alignment between these gradients as a proxy for contribution evaluation. We show that large-size gradient comparisons exhibit limitations in accurately assessing contributions, particularly in the context of deep neural networks. We make two key assumptions to address these limitations and refine the data valuation process. First, instead of evaluating the cosine similarity across all gradients or model parameters, we focus solely on the gradients of the last layer, providing computational advantages and a more accurate approximation of true Shapley values. Second, rather than just computing the marginal contributions, we calculate class-specific contributions, defining them as a measure of heterogeneity. Our contributions are as follows: 
\begin{itemize}
    \item We introduce a novel \textbf{contribution assessment} method (ShapFed) that precisely quantifies each participant's impact on the global model, capturing both their overall contribution and class-specific influence. 
    \item Building upon our contribution assessment approach, we propose a new \textbf{weighted aggregation} method (ShapFed-WA) that outperforms the conventional federated averaging algorithm (\cite{mcmahan2017communication}). 
    \item To enhance the collaborative fairness of the system, we \textbf{personalize} the updates sent from the server to clients based on their contributions. This ensures that clients making substantial contributions receive better updates than those with minimal or no contributions. 
\end{itemize}

\section{Literature Review} 

Data valuation is a well-studied topic in ML. The ``Data Shapley" framework proposed in \cite{ghorbani2019data} quantifies the value of individual training data points in model learning. This approach claims to offer more detailed insights into the significance of each data point, compared to traditional leave-one-out scores. Monte Carlo approximation methods have been proposed to estimate Shapley values (SVs) by sampling random permutations of data points and determining their marginal contributions. Further, \cite{jia2019towards} present algorithms designed to approximate the Shapley value with fewer model evaluations, thereby facilitating more efficient information sharing across different evaluations. These algorithms make specific assumptions about the utility function, including the stability of the learning algorithm and the characteristics of smooth loss functions.
Class-specific SVs were proposed in \cite{schoch2022cs} for more fine-grained data valuation. However, none of the above methods are applicable in the FL context because they require access to the raw data, which is not available at the orchestration server in FL.  

The work in \cite{sim2020collaborative} is the first to propose a collaborative learning scheme that considers incentives based on model rewards. They propose a data valuation method using information gain on model parameters given the data, by injecting Gaussian noise into aggregated data, to mitigate the need for a validation dataset agreed upon by all parties. Yet, even this method cannot be applied to FL directly because it assumes raw data sharing. Majority of the contribution assessment methods in FL require an auxiliary validation dataset, which introduces considerable time overhead in evaluating the contributions of participants. This issue is exemplified in the study by \cite{song2019profit}, where the authors introduce a novel metric known as ``contribution index''. This metric aims to assess the contribution of each data provider. The contribution index can be calculated using two proposed gradient-based methods (one round and multi rounds). These methods are designed to estimate confidence intervals, thereby providing a more refined and reliable measure of data contribution. However, the reliance on additional validation datasets and the time-intensive nature of these methods pose challenges to their practical implementation in real-world scenarios.

Gradient-based methods \cite{10.1145/3501811,xu2021gradient} have recently emerged as a practical approach for calculating Shapley values in FL. In \cite{10.1145/3501811}, the authors introduce the GTG-Shapley (guided truncation gradient Shapley) method. This approach combines between-round and within-round truncations to significantly reduce training costs. Between-round truncation is utilized to eliminate entire rounds of Shapley value calculations when the remaining marginal gain is small. Conversely, within-round truncation is applied to skip the evaluation of the remaining sub-models in permutations if the expected marginal gain is negligible, enhancing computational efficiency. 

Meanwhile, the study in \cite{xu2021gradient} proposed cosine gradient Shapley value (CGSV), which quantifies the contribution of a participant to the overall coalition based on the cosine similarity between the individual gradients and the aggregated gradient. This approach has been widely adopted in recent literature \cite{wu2023incentive,lin2023fair}. Furthermore, \cite{wang2020principled} proposed another dimension to SV approximation in FL: a sampling-based approximation and a group testing-based approximation. Works related to fairness in FL such as FedFAIM \cite{shi2022fedfaim} and FedCE \cite{jiang2023fair} also rely on gradient space to estimate the client contributions. Despite these advancements, the primary challenge lies in developing a strategy that effectively assesses contributions considering cost, practical feasibility, and per-class valuation of client data.

\section{Problem Formulation} 
In this work, the high-level goal is to solve the following standard cross-silo federated learning optimization problem: 
\begin{equation}
    \begin{split}
        f^{\star} \coloneqq & \min_{w \in \mathbb{R}^d} \Bigg[ f(w) \coloneqq \frac{1}{n} \sum_{i=1}^n f_i(w) \Bigg], \\ 
        \quad & f_i(w) \coloneqq \mathbb{E}_{\xi \sim \mathcal{D}_i}\big[F_i(w, \xi)\big], 
        \label{eq: optimization-problem}
    \end{split}
\end{equation}
where the components $f_i: \mathbb{R}^d \rightarrow \mathbb{R}$ are distributed among $n$ local participants and are expressed in a stochastic format $f_i(w) \coloneqq \mathbb{E}_{\xi \sim \mathcal{D}_i}\big[F_i(w, \xi)\big]$, where $\mathcal{D}_i$ represents the distribution of $\xi$ on participant $i \in \mathcal{N}$, where $\mathcal{N} = \{1, 2, \ldots, n\}$. This problem encapsulates standard empirical risk minimization as a particular case when each $\mathcal{D}_i$ consists of a finite number $m_i$ of elements $\{ \xi_1^i, \ldots, \xi_{m_i}^i \}$. In such cases, $f_i$ simplifies to $f_i(w) = \frac{1}{m_i} \sum_{j=1}^{m_i} F_i(w, \xi_j^i)$. Our approach does not impose any restrictive assumptions on the data distribution $\mathcal{D}_i$, and therefore our work covers the case of heterogeneous (non-i.i.d.) data where $\mathcal{D}_i \neq \mathcal{D}_j, \forall~ i \neq j$ and the \textit{local minima} $w_i^{\star} \coloneqq \arg\min_{w \in \mathbb{R}^d} f_i(w)$ might significantly differ from the global minimizer of the objective function (\ref{eq: optimization-problem}). 

\subsection{Preliminaries}
Let $\mathcal{M}_w: \mathcal{X} \rightarrow \mathcal{Y}$ be a supervised classifier parameterized by $w$, where $\mathcal{X} \subseteq \mathbb{R}^d$ and $\mathcal{Y} = \{1, 2, \ldots, M\}$ denote the input and label spaces, respectively, $d$ is the input dimensionality, and $M$ is the number of classes. We set $F_i(w, \xi) = \mathcal{L} (\mathcal{M}_w (x), y)$, where $\mathcal{L}$ is the loss function and $\xi\coloneqq(x,y)$ is a training sample such that $x \in \mathcal{X}$ and $y \in \mathcal{Y}$. \\

\paragraph{Shapley Values.} Shapley values, a concept derived from cooperative game theory, offer a principled approach to attributing the value of a coalition to its individual members \cite{winter2002shapley}. In the context of federated learning, Shapley values can be instrumental in quantifying the contribution of each participant to the learning of the global model. This is particularly relevant in FL, where diverse participants collaboratively train a model while keeping their data localized. For a federated learning system with $n$ participants, the Shapley value (contribution value) of the $i^{\text{th}}$ participant is defined as: 
\begin{equation}
    \resizebox{0.91\linewidth}{!}{
        $\phi_i(\nu) = \sum_{S \subseteq \mathcal{N} \setminus \{i\}} \frac{|S|! (n - |S| - 1)!}{n!} \Big( \nu\big(S \cup \{i\}\big) - \nu(S) \Big), $
    }
    \label{eq: shapley} 
\end{equation}
where $\mathcal{N}$ is the set of all participants, $S$ is a subset of participants excluding $i$, and $\nu(S)$ is the utility function that measures the performance of the subset $S$. The contribution estimation, $\phi_i(\nu)$, represents the average marginal contribution of participant $i$ over all possible coalitions. In FL, the utility function $\nu(S)$ can be defined as the performance of the global model trained using data from a subset of participants $S$. This can be measured in various ways, such as improvement in model accuracy or loss reduction. The challenge lies in the computational complexity of calculating Shapley values, as it requires evaluating the contribution of each participant across all possible subsets. To address this issue, we come up with an approximation method and describe it in Section \ref{sec: ca-approx}. 

\paragraph{Federated averaging.} A common approach to solving (\ref{eq: optimization-problem}) in the distributed setting is FedAvg \cite{mcmahan2017communication}. This algorithm involves the participants performing several local steps of SGD (local epochs) and communicating with the server over multiple communication rounds (i.e., every communication round consists of some local number of epochs). In each communication round, the updates from the participants are averaged on the server and sent back to all participants. For a local epoch $t$ and participant $i \in \mathcal{N}$, the local iterate is updated according to: 
\begin{equation}
    {w}^{t+1}_i = {w}_i^t - \eta_i^t \nabla F_i\big(w_i^t, \xi_i\big), 
\end{equation}
where $\xi_i = \{\xi_1^i, \ldots, \xi_{m_i}^i\}$. For a communication round, the update will be:
\begin{equation}
    w_s^{t+1} = \dfrac{1}{n} \sum_{i=1}^n \bigg(w_i^t - \eta_i^t \nabla F_i\big(w_i^t, \xi_i\big)\bigg).
\end{equation} 
The server then broadcasts the updated model $w_s^{t+1}$ to all participants, which is used as $w_i^{t+1}$ for the next local epoch. 

\begin{algorithm}[t]
    \caption{ShapFed algorithm}
    \renewcommand{\algorithmicrequire}{\textbf{input:}}
    \renewcommand{\algorithmicensure}{\textbf{procedure}}
    \begin{algorithmic}[1]
        \Require global weight initialization $w_s^1$, local learning rate $\eta$, no. of communication rounds $T$, no. of local epochs $K$, momentum factor $\mu$, $\mathcal{N} = [1, 2, \ldots, n]$ 
        \State Initialize $\gamma_i \gets {1}/{n}, \forall i \in \mathcal{N}$ 
        \For{$t = 1, \ldots, T$} 
            \If{$t = 1$} 
                \State Broadcast $w_s^t$ to participants: $\bar{w}_i^t \gets w_s^t, \forall i \in \mathcal{N}$ 
            \Else 
                \State Send $\bar{w}_i^t = \gamma_i w_s^t + (1 - \gamma_i) w_i^t$ to each party $i$ 
            \EndIf 
           
            \For{participant $i \in \mathcal{N}$ in parallel}
                \State Initialize local model ${w}_{i, 0}^t \gets \bar{{w}}_i^t$ 
                \For{$k = 1, \ldots, K$}
                    \State Sample $\xi_{i,k}^t$, compute $\nabla F_i \big(w_{i, k}^t, \xi_{i,k}^t\big)$ 
                    \State $w_{i, k+1}^t \gets w_{i, k}^t - \eta \nabla F_i\big(w_{i, k}^t\big)$ 
                \EndFor 
            \EndFor
            \State $w_s^{t+1} \gets \sum_{i=1}^n \tilde{\gamma}_i w_{i, K}^t$ 
            \State Compute $\tilde{\Gamma}$ using Equation \ref{eq: Gamma-cosine-sim}
            \State $\Gamma \gets \tilde{\Gamma}$ if $t=1$ else $\Gamma = \mu \Gamma + (1-\mu) \tilde{\Gamma}$ 
            \State Update $\gamma_i$ using Equation \ref{eq: importance-coef} 
        \EndFor
    \end{algorithmic}
\end{algorithm}

\section{Proposed Method} 
\subsection{Contribution Assessment: ShapFed} 
\label{sec: ca-approx}
Suppose that we define the utility function $\nu$ as class-wise performance assessed using a specific validation set. This framework enables the server to calculate \textit{Class-Specific Shapley Values} (CSSV), as delineated in Equation \ref{eq: shapley}. Since $\phi_i(\nu)$ is no longer a scalar, but a $M$-dimensional vector, we use the notation $\Gamma_i$ to denote CSSV. Although conceptually straightforward, this approach encounters practical challenges due to the necessity of a server-side validation set. This is problematic because users cannot share their data samples in FL due to privacy concerns. Furthermore, establishing a validation set that fairly evaluates all distributions $\mathcal{D}_i$, which could be extremely non-iid, presents a significant challenge. Some methods have tried to circumvent this by using publicly available datasets or creating synthetic samples \cite{li2023synthetic}. However, these solutions still require the server to perform a large number of inferences ($2^n-1$ calls) to compute CSSVs accurately. To overcome this, subsequent works \cite{xu2021gradient,jiang2023fair} have proposed using the directional alignment of gradients as an alternative utility measure.

\begin{figure}
    \centering
    \includegraphics[width=\linewidth]{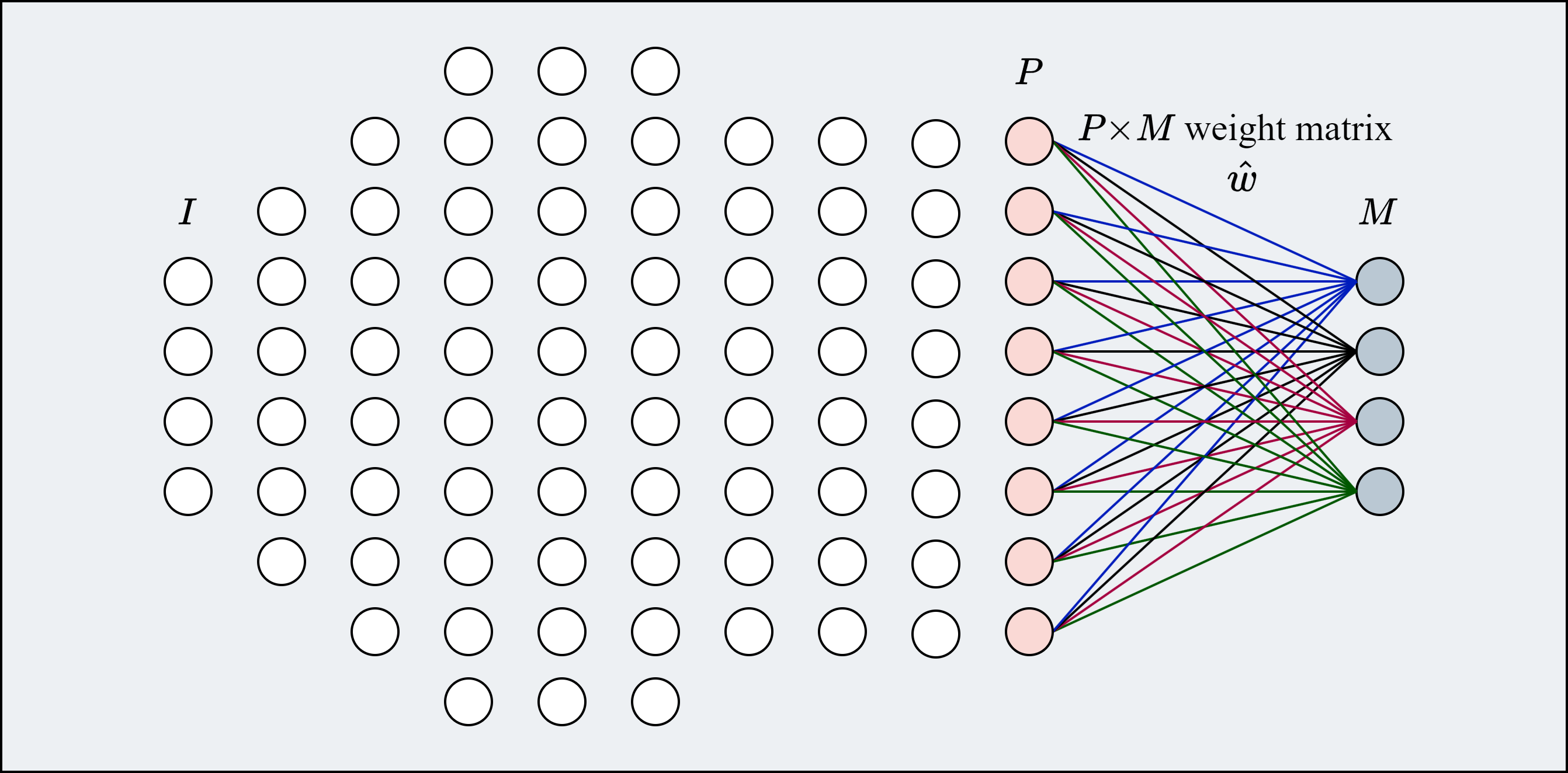}
    \caption{Illustration of the specific weight segments utilized for evaluating class-wise contributions and participant heterogeneity.}
    \label{fig: network-illustration}
\end{figure}

Motivated by this idea, we introduce an algorithm to evaluate CSSVs by utilizing the directional alignment between the gradients or network parameters of the last layer in the classifier $\mathcal{M}_w$. Conceptually, the classifier $\mathcal{M}_w$ can be viewed as a composition of a feature extractor that maps an input to a $P$-dimensional embedding ($\mathcal{X}\rightarrow\mathbb{R}^P$) and a linear classification layer (parameterized by $\hat{w} \in \mathbb{R}^{P \times M}$) that maps the feature embedding to a $M$-dimensional logits space ($\mathbb{R}^P\rightarrow\mathbb{R}^M$). Figure \ref{fig: network-illustration} illustrates an example network architecture, highlighting $\hat{w} $. We divide this matrix $\hat{w}$ into $M$ column vectors, with each vector corresponding to one of the output classes. 

In each communication round, the server collects the parameter updates or gradients ($w_i^t$) from all participants ($i \in \mathcal{N}$) and aggregates them to obtain $w_s^{t+1}$. For simplicity of notation, we drop the index $t$ in the subsequent discussion. Let $\hat{w}_i \subset w_i$ be the subset of updates corresponding to the last linear layer from participant $i$ and let $\hat{w}_s \subset w_s$ be the corresponding subset of the aggregated update computed by the server. Furthermore, let $\hat{w}_{i, j}$ ($\hat{w}_{s, j}$) denote the $j$-th column vector of $\hat{w}_{i}$ ($\hat{w}_{s}$), where $j \in \mathcal{Y}$. We define the contribution (CSSV) $\Gamma_i$ of participant $i$ as: 
\begin{equation}
    \begin{split}
        \Gamma_i \coloneqq \Big[\cos \big( \hat{w}_{i, 1}, \hat{w}_{s, 1} \big), & \cos \big( \hat{w}_{i, 2}, \hat{w}_{s, 2} \big), \ldots, \\ 
        & \cos \big( \hat{w}_{i, M}, \hat{w}_{s, M} \big)\Big]. 
    \label{eq: Gamma-cosine-sim} 
    \end{split}
\end{equation}

\noindent The $j$-th element of $\Gamma_i$ can be viewed as the contribution of participant $i$ ($i \in \mathcal{N}$) to the learning of the $j$-th class ($j \in \mathcal{Y}$), with higher values indicating stronger contribution. By the end of this process, the server obtains a matrix $\Gamma \coloneqq [\Gamma_1 , \Gamma_2, \ldots , \Gamma_n] \in [-1,1]^{n \times M}$, representing the heterogeneity and class-wise contribution of each participant. This matrix not only aids in understanding the individual contributions but also provides insights into the diverse nature of the data and learning across different classes. The CSSV value can be updated after every communication round with a momentum factor $\mu$ to facilitate smoother convergence.

\begin{figure}
    \centering
    \includegraphics[width=0.75\linewidth]{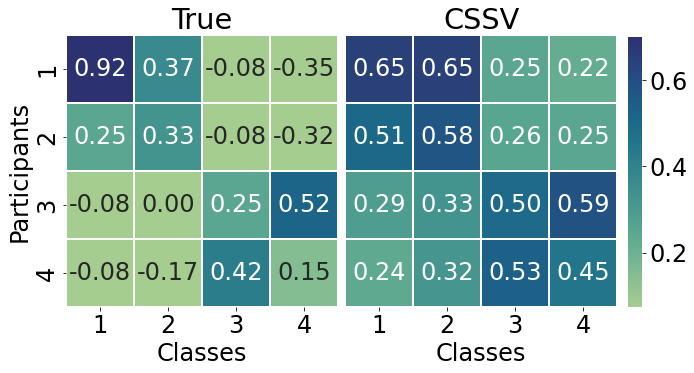} 
    \caption{Experimental demonstration comparing performance of utility functions $\nu$ on a synthetic dataset. The experiment contrasts the true Shapley value approach, using validation accuracy across all coalitions, with our proposed approximation method (Section \ref{sec: ca-approx}). }
    \label{fig: true-approx}
\end{figure}

\begin{example}[Identification of Class-specific Data Heterogeneity]
    Consider a scenario with $n=4$ and $M=4$ (as depicted in Figure \ref{fig: true-approx}), where the first two participants exclusively contain data belonging to the first two classes, and a similar splitting scenario applies to the remaining participants and classes. Ideally, our algorithm should generate block matrices in which the values of the diagonal blocks $\{\Gamma_{1:2, 1:2}, \Gamma_{3:4,3:4}\}$ are significantly higher than those in the off-diagonal blocks $\{\Gamma_{1:2, 3:4}, \Gamma_{3:4, 1:2}\}$. As shown in Figure \ref{fig: true-approx}, our contribution assessment method effectively discerns the heterogeneity among participants, offering a viable alternative to the resource-intensive computation of Shapley values. Additionally, our approach eliminates the need for a server-side validation set, a considerable advantage over methods that depend on such datasets. Finally, ShapFed requires only $\mathcal{O}(n + 1)$ inferences, a stark contrast to the $\mathcal{O}(2^n - 1)$ calls necessitated by exact SV computation (which becomes practically infeasible when $n \gg 0$.
\end{example}

It must be emphasized that the work of \cite{xu2021gradient} (CGSV) has already established that cosine similarity between gradients is a good approximation of SV (marginal contribution of each participant) and also provides a theoretical bound on this approximation error. ShapFed builds upon this result and \textit{extends CGSV to the class-specific setting}. The main difference with CGSV lies in how the overall SV is computed based on cosine similarity between gradients - while CGSV treats the complete model gradient as a whole and computes a single similarity value, we first estimate class-specific influences based on the gradients/weights of the last layer and average them, thereby providing a more granular understanding of the differences between the participants. 

\subsection{Weighted Aggregation: ShapFed-WA}
\label{section: weighted-aggregation}
While the FedAvg algorithm \cite{mcmahan2017communication} has become the standard approach in FL, it either treats all participants uniformly or weights them based on their self-reported training set size. However, this aggregation approach sometimes results in decreased accuracy \cite{kairouz2021advances}, particularly with non-iid (heterogeneous) data \cite{li2020federated,zhao2018federated} and when a large number of local epochs are employed \cite{kairouz2021advances}. The discrepancy between the local data distribution of each participant and the overall global distribution can cause the local objectives to diverge from the global optimum. Consequently, simple averaging may result in a sub-optimal solution or a slow convergence to the optimal solution.

\begin{figure}
    \centering
    \includegraphics[width=\linewidth]{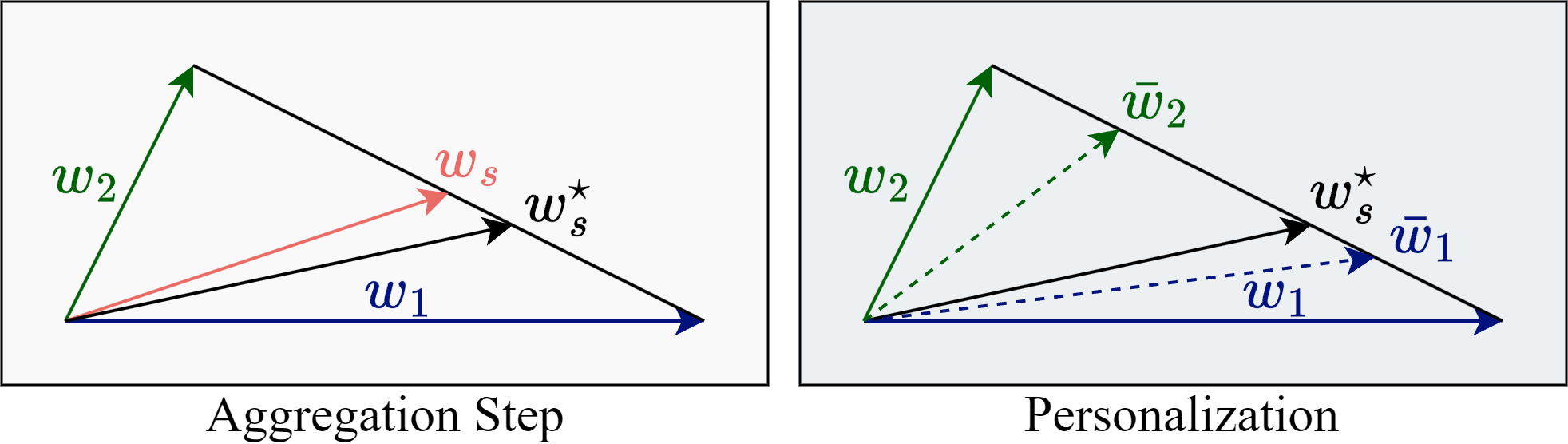} 
    \caption{{\textbf{Weighted Aggregation:}} The optimal weights $w_s^{\star}$ are derived using Equation \ref{eq: weighted-aggregation}, while {\color{red} $w_s$} represents the result of applying equal weights (FedAvg). {\textbf{Personalization:}} Rather than distributing a uniform global model to all users, we provide personalized weights $\bar{w}_i$, which are $\gamma_i$ combinations of individual user weights $w_i$ and the optimally aggregated weight $w_s^{\star}$. } 
    \label{fig:aggregation-personalization}
\end{figure}

To overcome this problem, we propose a weighted aggregation method called ShapFed-WA based on CSSV. Specifically, we first normalize CSSV between $0$ and $1$ and compute the weight assigned to each participant ($\gamma_i$) by averaging the normalized CSSV across all classes. We further normalize the participant weights such that they sum up to $1$. These operations are summarized in Equation \ref{eq: weighted-aggregation}. 

\begin{equation} 
    \gamma_i =  \frac{1}{M} \sum_{j=1}^M \left(\frac{1 + \Gamma_{i,j}}{2}\right), \quad  \tilde{\gamma_i} = \dfrac{\gamma_i}{\sum_{k=1}^n \gamma_k}.
    \label{eq: importance-coef} 
\end{equation} 
\noindent The final ShapFed-WA aggregation rule is given by: 
\begin{equation}
    w_s = \sum_{i=1}^n \tilde{\gamma}_i w_i.  
    \label{eq: weighted-aggregation} 
\end{equation} 

Figure \ref{fig:aggregation-personalization} shows a scenario where $w_s$ derived using standard FedAvg, assigning equal importance to each update ($w_s = 0.5 w_1 + 0.5 w_2$), leads to a sub-optimal outcome compared to homogeneous (iid) setting. In contrast, $w_s^{\star}$ represents the outcome of our proposed weighted aggregation method, which aligns more closely with the global optimum.

\subsection{Personalization}
Ensuring fairness within a FL system necessitates an incentive mechanism that appropriately rewards participants according to their contributions to the collaborative effort. In this work, we employ the following widely-used definition of collaborative fairness \cite{lyu2020collaborative,xu2021gradient,jiang2023fair}: \textit{In a federated setup, a high-contribution participant should be rewarded with a better-performing local model than a low-contribution participant. Mathematically, fairness can be quantified using Pearson's correlation coefficient between the standalone accuracies of participants and their respective final model accuracies.} 

To achieve collaborative fairness, we propose a personalization approach wherein the server transmits participant-specific updates instead of identical aggregated update. This technique is illustrated in Figure \ref{fig:aggregation-personalization} and can be calculated as: 
\begin{equation}
    \bar{w}_i = w_i + \gamma_i (w_s - w_i) = \gamma_i w_s + (1 - \gamma_i) w_i.
    \label{eq: personalization} 
\end{equation}

The above personalization method ensures that participants contributing less to the global model will progress towards the optimal solution $w^{\star}$ at a slower rate compared to those with higher contributions. Our strategy is akin to \cite{gasanov2022flix},  where the problem formulation involves an explicit mixture of local models and a global model (Equation 3 of \cite{gasanov2022flix}), and the mixture weight is a fixed hyperparameter. In contrast, we dynamically determine the mixture weights using our contribution assessment, which measures the quality of the updates received from each participant. Furthermore, the concept of personalization serves as a form of penalization in our method. Generally, low-contribution participants can detrimentally affect the performance of the global model, thereby potentially reducing the collaboration gain for high-contribution participants. Moreover, the low-contribution participants could be either free-riders, who seek to benefit from the global model without making any meaningful contribution, or even Byzantines, who intentionally send random updates without using any computational resources, thereby undermining the collaborative learning process. Such participants must be penalized with models having very low accuracy.

\section{Experiments and Results} 

\subsection{Datasets}
\textbf{CIFAR-10} \cite{cifar10}: This dataset comprises 60,000 RGB images, each with dimensions of 32 × 32 pixels, spanning 10 different classes. It is divided into a training set of 50,000 images and a testing set of 10,000 images. These images represent a diverse collection of objects, providing a comprehensive resource for evaluating image recognition models. \textbf{Chest X-Ray} \cite{tbchest}: The Tuberculosis (TB) Chest X-ray Database is a comprehensive collection of chest X-ray images containing 700 publicly accessible TB-positive images and 3500 normal images. \textbf{Fed-ISIC2019} \cite{ogier2022flamby}: This dataset is an amalgamation of the ISIC 2019 challenge dataset and the HAM1000 database, presenting a total of 23,247 dermatological images of skin lesions (8 classes). It encompasses data from six distinct centers, with respective data distributions of 9930/2483, 3163/791, 2961/672, 1807/452, 655/164, and 351/88. As a preprocessing step, we resized images to the same shorter side of 224 pixels while maintaining their aspect ratio, and normalized images’ brightness and contrast, as specified in \cite{ogier2022flamby}. 

\subsection{Experimental Setup}
\label{section:experimental-setup}
\paragraph{CIFAR-10.} We leverage the ResNet-34 architecture trained using the SGD optimizer with a fixed learning rate of $0.01$. For FL, we use 50 communication rounds. \paragraph{Chest X-Ray.} We employ a custom CNN architecture with three convolution layers followed by three fully connected layers. All the layers (except the last) are followed by ReLU activation. We use SGD optimizer with a learning rate of $0.01$, momentum of $0.9$, and weight decay of $5 \cdot 10^{-4}$. The models are trained for $50$ rounds. \paragraph{Fed-ISIC2019.} We employ the EfficientNet\_B0 model and use the same training settings as in CIFAR-10 with $200$ communication rounds. Due to the heterogeneity of this dataset, we use the balanced accuracy metric. 

Our experimental design involves dividing the training dataset among participants using two distinct strategies:
\begin{itemize}

\item \textbf{Imbalanced partitioning}: Here, the data is distributed unevenly among participants. We control the class distribution by manipulating two parameters: the major allocation probability and the number of classes designated as the majority group. In our experiments, we set the first class as the majority, with a higher allocation probability of 0.7, while the remaining classes are assigned equal probabilities of 0.1 each. This configuration allows us to investigate the model's behaviour in scenarios where data distribution is skewed. 

\item \textbf{Heterogeneous partitioning}: This strategy introduces a more complex and varied class allocation among participants. Each participant receives a unique distribution of probabilities across all classes. This scenario is designed to mimic real-world conditions where data distribution can be highly irregular and participant-specific. The Fed-ISIC2019 dataset fits into this purpose. For CIFAR-10, we implement label skew partitioning, where the first class is exclusively owned by the first participant, and the remaining nine classes are partitioned equally among all participants. For the Chest X-Ray, we adopt an equal major allocation strategy with a class variant among participants. Specifically, the first class is divided among 5 participants with probabilities 40\%, 30\%, 20\%, 10\%, and 0\%, respectively. The second class is similarly divided among 5 participants with probabilities 0\%, 10\%, 20\%, 30\%, and 40\%, respectively.

\end{itemize}

\subsection{Contribution Assessment} 
\begin{figure}[t] 
    \centering 
    \includegraphics[width=0.65\linewidth]{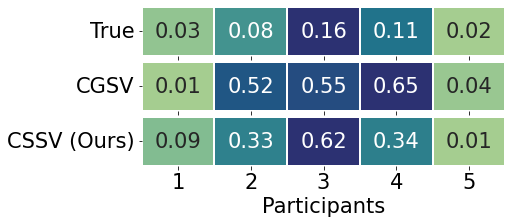}
    \caption{Comparison of our proposed contribution assessment algorithm (CSSV) with CGSV and true Shapley value computations using ResNet-34 architecture on Chest X-Ray dataset. }
    \label{fig:chestxray-cssv} 
\end{figure} 

\begin{figure}[t] 
    \centering
    \includegraphics[width=\linewidth]{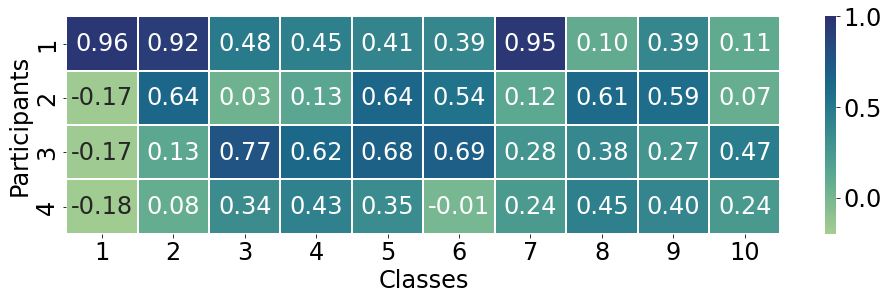}
    \caption{Heatmap visualization of class-specific Shapley values for heterogeneous setting (explained in Section \ref{section:experimental-setup}) evaluated on CIFAR-10 dataset.}
    \label{fig:cifar10-imb}
\end{figure}

Figures \ref{fig:chestxray-cssv} and \ref{fig:cifar10-imb} provide heatmap visualizations of CSSVs for Chest X-Ray and CIFAR-10 datasets, respectively, using ResNet-34 architecture as outlined in Section \ref{section:experimental-setup}. 
\begin{itemize}
    \item Figure \ref{fig:chestxray-cssv} offers a comparative analysis of the true estimation of Shapley values (with the utility function being validation accuracy on an auxiliary set), CGSV, and our proposed contribution assessment approach, CSSV. In the given data distribution scenario, where each participant holds equal data volumes, $P_3$ is expected to have a higher utility and contribution measure. Conversely, $P_1$ and $P_5$ are predicted to score the lowest in terms of contribution, with $P_2$ and $P_4$ expected to have moderate contributions. The obtained empirical results align with these expectations for rows 1 and 3 (true estimation and our approach), demonstrating the accuracy of our method. However, the CGSV approach deviates by assigning a disproportionately high score to $P_4$ and comparable scores to $P_2$ and $P_5$. This discrepancy stems from CGSV's reliance on all network layers for Shapley value computation, which becomes a limitation in the context of large models such as ResNet-34. Our method, on the other hand, effectively identifies the importance of each participant and closely tracks the true Shapley value while being computationally efficient. 
    \item Figure \ref{fig:cifar10-imb} showcases a heatmap representing contributions in the CIFAR-10 dataset under a heterogeneous setting. Here, $P_1$ possesses all entries for class 1, while the remaining classes are evenly distributed among all four participants. Consistent with expectations, participants $P_2$, $P_3$ and $P_4$ exhibit the lowest contribution scores for class 1. This outcome highlights the effectiveness of our approach in identifying and quantifying participant contributions in a scenario where data is heterogeneous. 
\end{itemize}

\begin{figure}[H] 
    \centering
    \includegraphics[width=\linewidth]{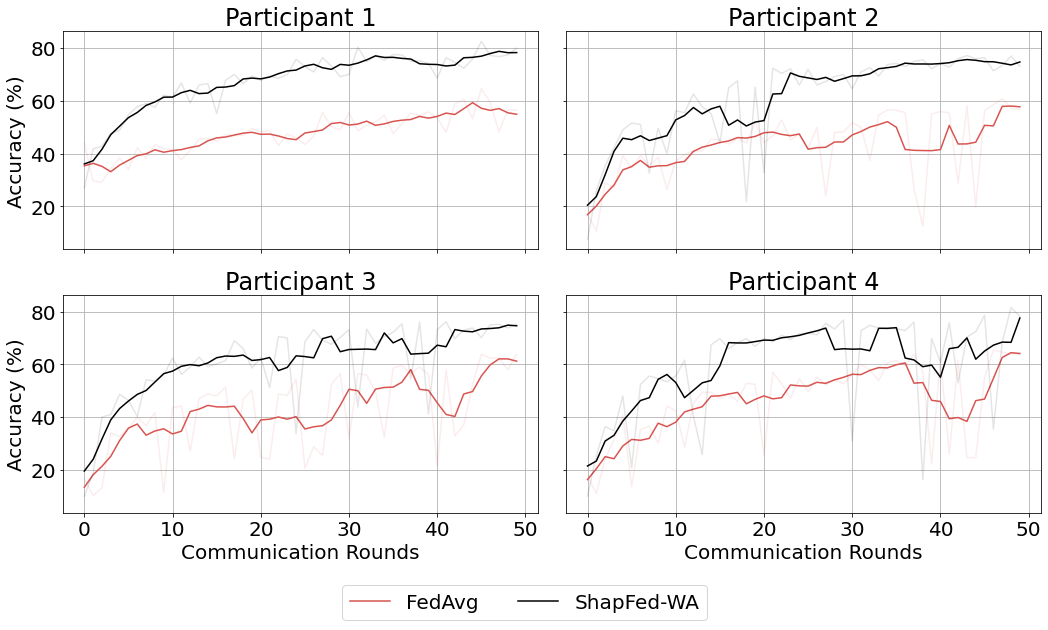} 
    \caption{Comparing FedAvg and ShapFed-WA on CIFAR10 under an imbalanced split scenario: insights into the balanced accuracy of four individual participants.} 
    \label{fig:weightComparison} 
\end{figure}

\subsection{Weighted Aggregation}
\paragraph{CIFAR-10.} We evaluate the performance of  ShapFed-WA and FedAvg on the CIFAR-10 dataset across the two split scenarios. The results in Figure \ref{fig:weightComparison} demonstrate the superiority of our method over {FedAvg} in terms of (balanced) validation accuracy for both the global model and individual participant models (in imbalanced partitioning). 

\begin{figure}[t]
    \centering
    \includegraphics[width=\linewidth]{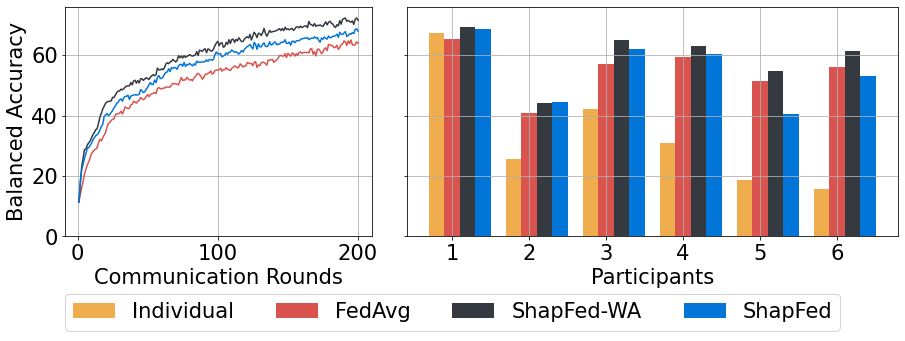}
    \caption{(\textbf{Left}) The balanced accuracy of our methods (ShapFed-WA \& ShapFed) vs FedAvg. (\textbf{Right}) Per-participant accuracy using all methods evaluated on Fed-ISIC2019 dataset.} 
    \label{fedisiccompare}
\end{figure}

\paragraph{Fed-ISIC2019.} Our method consistently outperformed the FedAvg method as shown in Figure \ref{fedisiccompare}. This improvement can be attributed to ShapFed-WA's ability to dynamically adjust based on the varying contributions of individual clients. Comparative effectiveness of our approach in terms of participant-wise balanced accuracy is clearly illustrated in histogram plot, in Figure \ref{fedisiccompare}.

\begin{table}[b] 
    \centering
    \resizebox{\linewidth}{!}{ 
        \begin{tabular}{l|l|l
            |S[table-format=2.2, detect-weight] 
            S[table-format=2.2, detect-weight] 
            S[table-format=2.2, detect-weight] 
            S[table-format=2.2, detect-weight] 
            S[table-format=2.2, detect-weight] 
            |S[table-format=2.3, detect-weight] 
        } 
            \toprule 
            \rowcolor{lightblue} \multicolumn{2}{c|}{\textbf{Dataset} / \textbf{Partition}} & \textbf{Setting} & $\vect{P_1}$ & $\vect{P_2}$ & $\vect{P_3}$ & $\vect{P_4}$ & $\vect{P_5}$ & \textbf{Corr.} \\ \midrule \midrule  
            \multirow{3}{*}{\rotatebox[origin=c]{0}{ChestXRay}} & \multirow{3}{*}{\rotatebox[origin=c]{0}{Het.}} & Individual & 50.0 & 64.7 & 62.0 & 53.7 & 50.0 & $\mbox{---}$ \\ 
            & & FedAvg & 50.0 & 55.8 & 61.9 & 54.2 & 50.0 & 0.82 \\ 
            & & \cellcolor{lightgreen}ShapFed & \cellcolor{lightgreen}50.0 & \cellcolor{lightgreen}65.2 & \cellcolor{lightgreen}69.5 & \cellcolor{lightgreen}58.5 & \cellcolor{lightgreen}50.0 & \cellcolor{lightgreen} \bfseries 0.93 \\ \midrule 
            \multirow{8}{*}{\rotatebox[origin=c]{0}{CIFAR-10}} & \multirow{4}{*}{\rotatebox[origin=c]{0}{Imb.}} & Individual & 75.8 & 45.4 & 48.6 & 31.6 & $\mbox{---}$ & $\mbox{---}$ \\ 
            & & FedAvg & \cellcolor{lightred} 56.6 & 56.8 & 63.8 & 64.2  & $\mbox{---}$ & -0.60 \\

          \ &  & CGSV &  \cellcolor{lightred} 57.2  & 59.0 & 58.8  &  60.4  &  $\mbox{---}$ & -0.98\\

            & & \cellcolor{lightgreen}ShapFed & \cellcolor{lightgreen} 81.4 & \cellcolor{lightgreen} 78.2 & \cellcolor{lightgreen} 71.8 & \cellcolor{lightgreen} 73.6  & \cellcolor{lightgreen} $\mbox{---}$ & \cellcolor{lightgreen} \bfseries 0.74\\ \cmidrule{2-9} 
            
            & \multirow{4}{*}{\rotatebox[origin=c]{0}{Het.}} & Individual & 75.2 & 68.8& 66.8& 69.0 & $\mbox{---}$ & $\mbox{---}$ \\ 
            & & FedAvg & \cellcolor{lightred} 74.6 &70.2 & 70.2 & 76.0 & $\mbox{---}$ & 0.53\\ 
                        &  & CGSV &   \cellcolor{lightred} 55.0 & \cellcolor{lightred} 55.8 & \cellcolor{lightred} 57.2  &  \cellcolor{lightred} 52.6  &  $\mbox{---}$ &  -0.26\\ 
            
            & &\cellcolor{lightgreen}ShapFed & \cellcolor{lightgreen} 79.8& \cellcolor{lightgreen} 75.4 & \cellcolor{lightgreen} 69.0 & \cellcolor{lightgreen} 75.0  & \cellcolor{lightgreen} $\mbox{---}$ & \cellcolor{lightgreen} \bfseries 0.90\\ \bottomrule 
        \end{tabular} 
    } 
    \caption{Performance and fairness comparison with our method and FedAvg. We use Pearson's correlation ($\uparrow$) as a fairness metric on CIFAR-10. The red highlight indicates a negative gain from collaboration.} 
    \label{broadcasting} 
\end{table}

\subsection{Personalization}
We assessed the personalization approach by observing individual participant validation accuracy under three distinct training conditions: first, when each participant trained exclusively on its data, second, when trained collaboratively with other participants using our methods: ShapFed-WA and ShapFed, and third when trained collaboratively with other participants using FedAvg method. We also report the collaborative fairness (Pearson's correlation coefficient) of these methods. 

\paragraph{CIFAR-10.} In Table \ref{broadcasting}, we report the balanced accuracy for Individual, FedAvg, CGSV and ShapFed approaches for each participant. In imbalanced partitioning, faster convergence was observed for the first participant model, which held a significant 70\% of the data. Conversely, participants with lower contributions exhibited slower progress towards the optimal solution. This implies that the participants receive distinct updates based on their contributions to the collaborative learning process. On the other hand, models that are trained using the FedAvg and CGSV algorithms, exhibit close accuracy, implying unfair treatment. CGSV demonstrated a negative correlation - lower collaborative fairness of $-0.98$. A similar pattern is observed in heterogeneous partitioning, where higher correlation values with individual accuracies imply a fairer consideration of diverse client contributions. 

\paragraph{Chest X-Ray.} As anticipated, our proposed ShapFed outperformed FedAvg in this scenario. Additionally, even though the five participants have equal amounts of data split among them, their class distributions vary. Notably, the third participant experienced the most substantial gain from the collaboration, resulting in higher accuracy.  \\ 

\paragraph{Fed-ISIC2019.} 

As shown in Table \ref{corr}, our method exhibits a strong correlation (0.84) with individual client performances, indicating a high degree of fairness. In contrast, FedAvg demonstrates a significantly lower correlation (0.63), implying a more uniform distribution of updates regardless of individual client contributions. As such, FedAvg has a negative collaboration gain for $P_1$ (highlighted in red). 
These findings confirm that our methods effectively recognize and incorporate the distinct contributions of each client in a collaborative learning environment with complex data distributions.

\begin{table}[t!]
\centering
\resizebox{\linewidth}{!}{
    \begin{tabular}{
        l
        |S[table-format=2.2, detect-weight]
        S[table-format=2.2, detect-weight]
        S[table-format=2.2, detect-weight]
        S[table-format=2.2, detect-weight]
        S[table-format=2.2, detect-weight]
        S[table-format=2.2, detect-weight]
        |S[table-format=1.3, detect-weight]
        }
            \toprule 
            \rowcolor{lightblue} \textbf{Setting} & $\vect{P_1}$ & $\vect{P_2}$ & $\vect{P_3}$ & $\vect{P_4}$ & $\vect{P_5}$ & $\vect{P_6}$ & \textbf{Corr.} \\ \midrule \midrule 
            Individual & 67.2   & 25.7  & 42.3  & 31.0  & 18.5  & 15.6  & $\mbox{---}$ \\ 
            FedAvg & \cellcolor{lightred} 65.4 & 40.9  & 57.2  & 59.3  & 51.5  & 56.2  & 0.63 \\ \midrule 
            \rowcolor{lightgreen} ShapFed-WA  & 69.3 & 44.3  & 65.0  & 63.1  & 54.8  & 61.2  & 0.62 \\ 
            \rowcolor{lightgreen} ShapFed & 68.5 & 44.4  & 61.9  & 60.4  & 40.6  & 53.2  & \bfseries 0.84 \\ \bottomrule 
    \end{tabular}
}
    \caption{Performance and fairness comparison using Pearson's correlation ($\uparrow$) as a fairness metric on Fed-ISIC2019. The red highlight indicates a negative gain from collaboration. }
\label{corr}
\end{table}

\section{Summary} 
This work proposes Class-Specific Shapley Values (CSSVs) to quantify participant contributions at a granular level. The contributions of this work include a novel method to deepen the understanding of participant impact and improve fairness analysis. Evaluation against FedAvg shows superior performance and additional experiments reveal enhanced fairness by personalizing client updates based on contributions. Overall, the approach aims to achieve a more equitable distribution of benefits in FL. In future, we plan to conduct an in-depth theoretical analysis aimed at identifying the specific characteristics that contribute to an effective estimation of Shapley values. This analysis will enhance our understanding of the factors that influence the accuracy and reliability of Shapley value approximations. Furthermore, an investigation into what makes our approximation of cosine similarity from the last layer a robust indicator of contributions will be explored.

\bibliographystyle{named}
\bibliography{ijcai24}

\newpage 
\appendix 
\section{Contribution Assessment Results}
\begin{figure}[!htbp] 
    \centering
    \includegraphics[width=\linewidth]{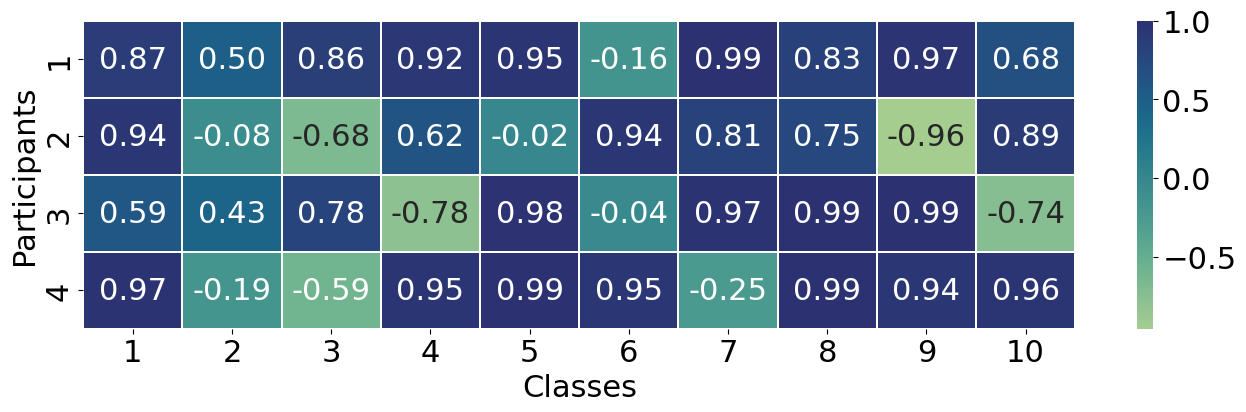}
    \caption{Heatmap visualization of class-specific shapley values for imbalanced setting evaluated on CIFAR-10 dataset.}
    \label{fig:cifar10-het} 
\end{figure}

In this section, 
we aim to further enhance understanding of our contribution assessment. The heatmap plot, illustrated in Figure~\ref{fig:cifar10-het}, presents the contributions made across the CIFAR-10 dataset under an imbalanced data splitting scenario, where participant 1 holds a significant portion of the data. As expected, participant 1 emerges as the primary contributor across all classes. While other participants do make substantial contributions to specific classes, participant 1 remains the most influential contributor. 

\section{Weighted Aggregation Results}
\begin{figure}[!htbp] 
    \centering
    \includegraphics[width=\linewidth]{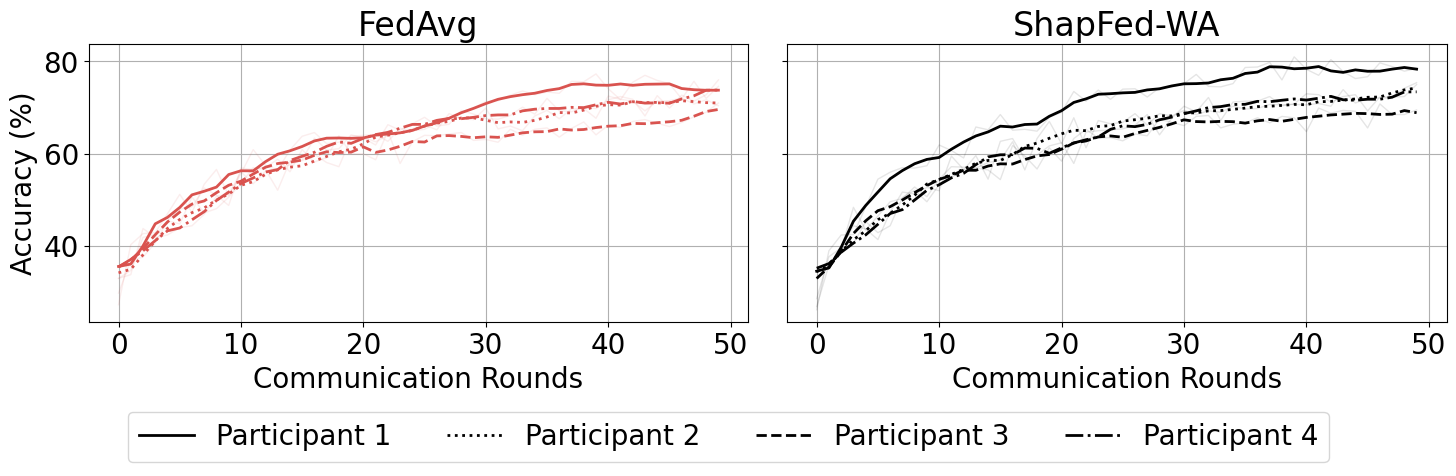} 
    \caption{Comparing FedAvg and ShapFed-WA on CIFAR10 under the heterogeneous splitting scenario: insights into the balanced accuracy of four individual participants.} 
    \label{fig:weightComparisonCifar} 
\end{figure}

\begin{figure}[t] 
    \centering
    \includegraphics[width=\linewidth]{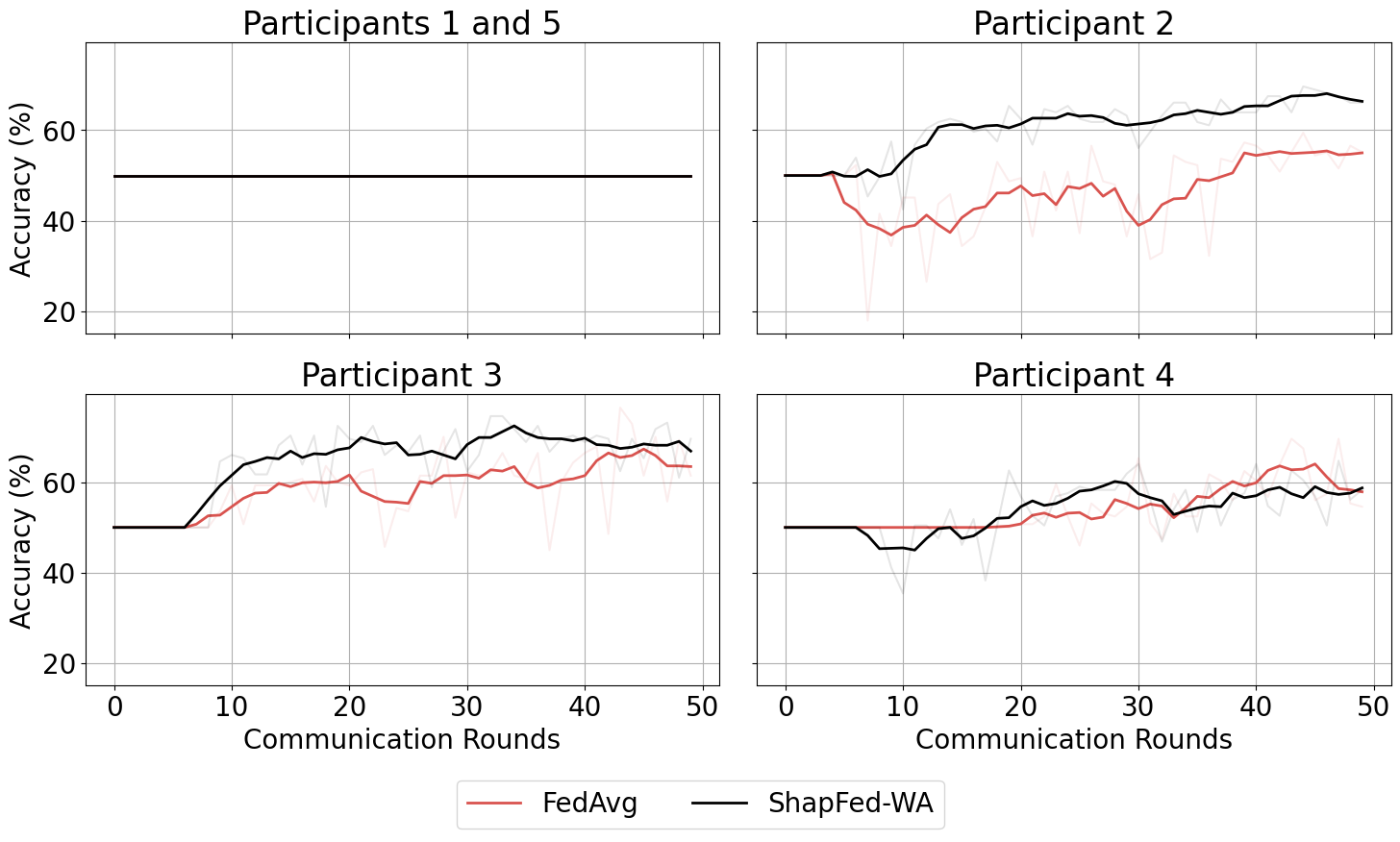} 
    \caption{Comparing FedAvg and ShapFed-WA on Chest X-Ray under the heterogeneous splitting scenario: insights into the balanced accuracy of five individual participants.} 
    \label{fig:weightComparisonChest} 
\end{figure}

To enhance the evaluation of integrating weighted aggregation in optimizing federated learning workflows, we explore the performance of federated learning methods using CIFAR-10 and Chest X-Ray datasets with heterogeneous data distribution. Our analysis primarily contrasts the proposed ShapFed-WA method against the traditional FedAvg algorithm. The results, depicted in Figures \ref{fig:weightComparisonCifar} and \ref{fig:weightComparisonChest}, highlight the superior performance of our approach compared to FedAvg, in terms of both the overall accuracy of the global model and the balanced accuracy of individual participant models. In Figure~\ref{fig:weightComparisonCifar}, participant 1, who holds a larger data share, demonstrates higher accuracy. \footnote{The accuracies of each participant, prior to aggregation on the server, are illustrated.}  

\section{Fed-ISIC2019 Results} 

\begin{figure}[t] 
    \centering
    \includegraphics[width=0.775\linewidth]{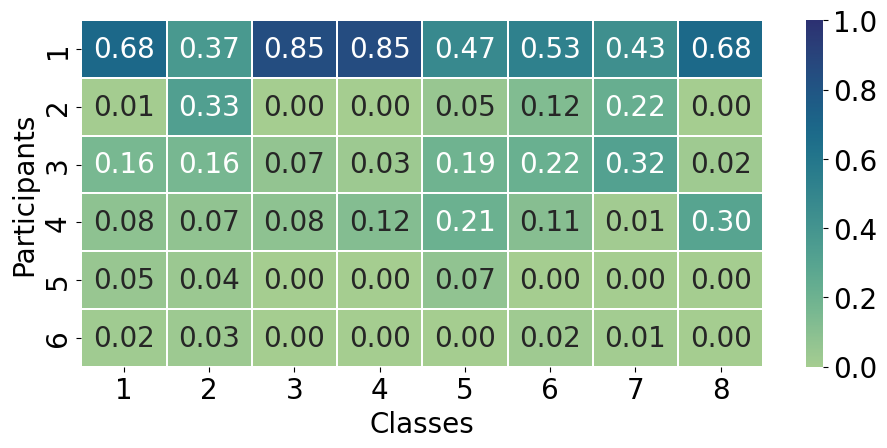}
    \caption{Heatmap visualization of class distributions in the Fed-ISIC2019 dataset.} 
    \label{fig:fedisicdist}
    \centering
    \includegraphics[width=0.775\linewidth]{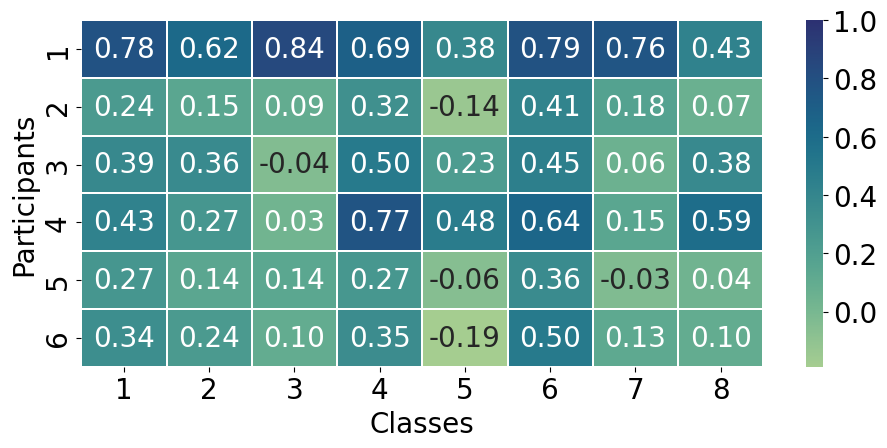} 
    \caption{Heatmap visualization of class-specific shapley values ($\Gamma$) for participants in Fed-ISIC2019 dataset.}
    \label{fig:fed}
\end{figure}

This analysis examines the class distribution among participants in a federated learning setting and the calculated contribution values of each participant (CSSVs), as shown in Figures~\ref{fig:fedisicdist} and \ref{fig:fed}, respectively. The heatmap plot in Figure~\ref{fig:fedisicdist} displays the proportion of the data points each participant holds per class. Our key findings are as follows: 
\begin{itemize}
    \item Participant 1 consistently achieves higher Shapley values (SVs) across most classes, which aligns with expectations given their larger share of data points. 
    \item Participants 2-6 exhibit generally lower or comparable SVs, with Participant 6 outperforming Participant 5 in average per-class SVs. This underscores the method's emphasis on data quality over quantity.
    \item Despite having smaller data volumes, Participants 3 and 4 attain higher per-class SVs than Participant 2, reflecting the valuable knowledge these participants contribute across all classes. 
\end{itemize} 

\noindent These insights illustrate the improved fairness that ShapFed introduces to federated learning.

\end{document}